\documentclass{article}

\usepackage{arxiv}

\usepackage[utf8]{inputenc} % allow utf-8 input
\usepackage[T1]{fontenc}    % use 8-bit T1 fonts
\usepackage{hyperref}       % hyperlinks
\usepackage{url}            % simple URL typesetting
\usepackage{booktabs}       % professional-quality tables
\usepackage{amsfonts}       % blackboard math symbols
\usepackage{nicefrac}       % compact symbols for 1/2, etc.
\usepackage{microtype}      % microtypography
\usepackage{lipsum}
\usepackage{amsthm}
\usepackage{amsmath}
\usepackage{graphicx}
\usepackage{subfigure}
\usepackage{amssymb}

\title{"Drunk Man" Saves Our Lives: Route Planning by a Biased Random Walk Model}

\author{
  Xinyi Hu\thanks{Work done in East China Normal University, Department of Mathematical Sciences.} \\
  Donald Bren School of Information and Computer Sciences\\
  University of California, Irvine\\
  Irvine, CA 92697 \\
  \texttt{xinyih20@uci.edu} \\
   \And
 Quchen Miao\\
  School of Physics and Electronic Science\\
  East China Normal University\\
  Shanghai, China \\
  \texttt{qcmiao@stu.ecnu.edu.cn} \\
   \And
  Zexuan Zhao\\
  Department of Biology\\
  University of Maryland, College Park\\
  College Park, MD 20742 \\
  \texttt{zx.zhao19980417@gmail.com} 
}

\begin{document}
\maketitle

\begin{abstract}
Based on the hurricane struking Puerto Rico in 2017, we developed a transportable disaster response system "DroneGo" featuring a drone fleet capable of delivering medical package and videoing roads. Covering with genetic algorithm and a biased random walk model mimicing a drunk man to explore feasible routes on a field with altitude and road information. A proposal mechanism guaranteeing stochasticity and an objective function biasing randomness are combined. The results shown high performance though time-consuming.
\end{abstract}

% keywords can be removed

\section{Introduction}
Based on the hurricane struking Puerto Rico in 2017, we developed a transportable disaster response system "DroneGo" featuring a drone fleet capable of delivering medical package and videoing roads. Assuming equal weight for both mission, we take the capability of carrying out the former missions as a constraint and a starting point from which reconnaissance routes are built. 

The feasibility of fitting packages into cargo bay 1 or 2 is tested by genetic algorithm. In scenario where drones carry packages to and unloaded back, from specification of drones and loading weight can we derive the maximum reachable distance of each drone loaded. A k-means clustering algorithm is used for partitioning destinations and deriving centroids as locations of bases. Sampled points from roads are added to take into account reconnaissance mission. Points of destination are oversampled with increasing weight till centroids fall within the scope of the maximum reachable distance. Drones for each base are selected based on their performance and medical demand of destinations. Containers of each base is filled till full with minimum drones required and maximum medicine storage. Transportation of medical packages and backup drones are also included in our system.

A biased random walk model mimicing a drunk man is built to explore feasible routes on a field with altitude and road information. A proposal mechanism guaranteeing stochasticity and an objective function biasing randomness are combined. Road and home attraction are weighted differently as walk goes, to simulate the behavior of an exploratory and nostalgic drone. In modified model, we unleashed nuisance parameters to follow a log nomal distribution and use recurrent neural network to learn from previous routes and enhance its performance. The results of modified model are apparantly better and a conbination of two routes is chosen for our system. When analyzing filtered distribution of nuisance parameters and their contribution to performance, we find them neither contributing or providing reasonable experience for us to pick their values in the next run. The performance difference caused by k-nearest neighbor rule selection is huge, but the underlying mechanism is not explorable in this single field.

\begin{table}[p]
  \caption{Symbol Description} 
  \label{notations}
  \centering
  \begin{tabular}{cc}
    \toprule
		Symbol & Description\\
		\midrule
		MPC & max Payload capacibilities of drones\\
		MDC & max drivable capacibility of drones\\
		k & the proportional coefficient for MDC/MPC\\
		l & the weight of load\\
		t & maximum flight time with cargo\\
		T & maximum flight time without cargo\\
		$P_1$ & the power of flying to destination\\
		$P_2$ & the power of flying back\\
		$t_1$ & the time of flying to destination\\
		$t_2$ & the time of flying back\\
		$x_1,x_2,\cdots,x_n$ & the longitude of points to be clustered in K-means clustering\\
		$y_1,y_2,\cdots,y_n$ & the latitude of points to be clustered in K-means clustering\\
		dist(X,Y) & the Euclidean distance between X and Y\\
		$label_i$ & point that has the least squared Euclidean distance\\
		$a_i$ & the centroid of each cluster\\
		$p_{road}$ & the tendency of drones flying towards roads\\
		$p_{home}$ & the tendency of drones going home\\
		$\gamma$ & relative weight of $p_{home}$\\
		$\hat{p}_{road}$ & enlarged difference of $p_{road}$\\
		$\hat{p}_{home}$ & enlarged difference of $p_{home}$\\
		$\alpha$ & constant relevant to $\gamma$\\
		$\beta$ & constant relevant to $\gamma$\\
		MFD & maximum flight distance\\
		d & distance from current cell to the origin\\
		CNN & convolutional neural network\\
		RNN & recurrent neural network\\
		t-1,t,t+1 & a Time-Series data set\\
		X & input data set\\
		W & input weight\\
		U & weight of sample in this time\\
		V & output weight\\
		CCR & conbinational coverage rate\\
		NC & net coverage\\
		A & the area of a specified bounding box\\
		$\hat{f}(x)$ & indicates the distribution function of after filtering\\
		$f(x)$ & indicates the distribution function of before filtering\\
		$m(\hat{f}(x))$ & measure of $\hat{f}(x)$. from measure theory to probability theory.\\
		$m(f(x))$ & measure of $f(x)$, from measure theory to probability theory.\\
		\bottomrule
  \end{tabular}
\end{table}

\section{Introduction}
\subsection{Symbol Description and Disambiguity}
\subsubsection{Symbol Description}

For symbol description, see table \ref{notations}.

\subsubsection{Disambiguity}

\begin{enumerate}
  \item A base is where container is located and drones depart. Bases store medicine and retrieve drones to process videos for decision-making.
  \item A destination is where medical package is delivered to.
  \item A package plan of destination A is all the medical packages needed by destionation A in a day.\label{package plan}
  \item Word "feasible" has different meanings: feasible base means drones that depart from it could reach the destination; feasible drone plan means the packages can be packed into the drone cargo bay; feasible cell means drone can walked on it for reason defined in context; feasible route means the drone goes back home in the end.
  \item Field: The place where our drone walks on. It has all the factors perceptible and necessary to our drone.
  \item Road coverage: the sum of the importance of cells on the route, used for route evaluation.
\end{enumerate}

\subsection{Problem Restatement}

The occurance of natural disaster events recorded has been increasing exponetially since 1900, despite the recent minor decline \cite{owidnaturaldisasters}. The horrific casualty and economic damage caused by catastrophes concerns every Earth citizen who dreams about living a safe and comfortable life. Not only does timely and adequete response to natural disasters helps decrease casualty and property loss, but also it reassures people from anxiety in the context of global climate change.

Required by HELP, Inc., we developed a transportable disaster response system, "DroneGo", based on the hurricane that struck Puerto Rico in 2017. It includes a drone fleet and medicine configurated according to anticipated medical package demand of destination. It's carried in ISO standard cargo containers and transported to elaborately picked bases from which drones depart with two missions --- medical package delivery and video recording. 

\subsection{Our Work}

We divided the ultimate goal into two sub-problems:
\begin{itemize}
  \item Selecting base locations and drones and making delivery plan.
  \item Making reconnaissance routes. 
\end{itemize}

In the first problem, locations are given by k-means clustering of oversampled delivery locations and sampled points on main roads. The maximum reachable distance of different drones are calculated in the scenario that drones carry packages when flying to the delivery locations and are unloaded when going back. The maximum flight time under specific loading weight is derived from a linear model according to max payload capability and flight time with no cargo given by attachment 2. All combination of drones and package plans \ref{package plan} are numerated and tested by a genetic algorithm for whether they can be packed into the drone cargo bay. The flight capabilities of drones serve as constraints to our k-means clustering in a way that the oversampling weight of delivery points increases till each cluster centers fall within the overlapping scope of maximum reachable circle of destinations. The most requirement-satisfactory drones are selected based on the performance and missions assigned.

In the second problem, we prototyped initial feasible routes by simulating a biased random walk model on a raster featuring altitude and importance derived from the place main roads are located. A proposal mechanism is used to move drones from current cell to a randomly-picked adjacent available cell. Drones decide whether to accpet or reject the proposal based on an objective function of importance and distance from the starting cell. In optimization stage, we integrated feasible and well-performed routes into the objective function, weaken bias by recurrent neural network and unleashed hard-coded parameters by specifying them to follow a lognormal distribution. We also combined 2 routes to get higher coverage to elevate the reconnaissance ability of our system. 

\subsection{Assumptions}
\begin{enumerate}
  \item When departing for delivery mission, drones first rise to the height required for the route; When landing, drones move right above the spot then go straight down. Time for going straight up and down is negligible.
  \item Drones carry out missions seperately.
  \item When filming, drones keep at a specific height above ground to keep the resolution of videos.
  \item The maximum distance from which roads can be filmed is $ 50m $\label{50m}.
  \item Only bases have charging facilities for drones.
  \item One flight has only one destination and the drone carries all the medical packages the destination needs for a day.
  \item Maximun flight time decreases linearly with the total weight of cargos with coefficient $k \leq 1.3$\label{linear decrease}.
  \item Drones' speed is constant, regardless to cargo weight.\label{speed consistancy}
  \item The weight of buffering material can be ignored.
  \item Medicine would not go bad, therefore we should fill the container with as many medical packages as possible.
\end{enumerate}

\section{Data Manipulation}
\subsection{Data Source}

We obtained a georeferenced raster image displaying elevation data for Puerto Rico and the U.S. Virgin Islands derived from NED data released in December, 2010\cite{altitudedata}. The data is in $100\, m \times 100\,m$ resolution and uses the Albers Equal-Area Conic projection. 

Road shape data is from OpenStreetMap, a open data licensed under the Open Data Commons Open Database License (ODbL) by the OpenStreetMap Foundation (OSMF)\cite{OpenStreetMap}.
Thanks to thousands of individuals who contributed to the database.

\subsection{Data Wrangling}

We cropped the altitude raster into a $ 700 \times 1900 $ extent, covering the entire mainland of Puerto Rico. 
World Geodetic System (WGS84) is the coordinate reference system to which the raster is projected to acquire the longitude and latitude of cells.

We filtered out minor roads in shape files to include only motorways, divided roads, national roads and regional roads. To transform shape files to one single raster for our drones to walk on, we buffered lines with $ 50\,m $ radius\ref{50m} and masked a template raster with derived polygons. For each road types, we assigned a progressively decreasing series $(2, \sqrt{3}, \sqrt{2}, 1)$ as their importance and picked the highest importance if one cell belongs to more than once type.\label{importance}

To integrate altitude and road data, we stacked them to a brick file with the same extent and CRS as one of the input data of our model.

\section{Part1: Container Configuration}
\subsection{Analysis of the problem}

We came up with three considerations when choosing the best location for bases:
\begin{itemize}
  \item Drones should be able to fly from the base to distination with medicine and come back unloaded;
  \item Drones can fly from one base to another to make up for medicine shortage.
  \item The number of roads that drones can monitor, measured by the number of cells of roads;
\end{itemize}

All three factors involves calculation of maximum flight distance of drones carrying medical packages, which should be thought about first. Because we don't consider time and energy consumption, if the base is feasible for delivering medicines, the actual distance between the base and destination is ignored when scheduled. In reality, this could be compensated by forcing drones to deliver packages at night because daytime is more valuable for reconnaissance. 

After bases are determined, drones could be selected according to performance and the medicine demand of distination. When filling containers, we first decided the type and number of drones. Then the medical packages
are filling into the rest space of the container in a given proportion till the container is full.

Due to the fact that one base may have too many destinations, and thus medical packages are consumed much faster than other bases, we considered making it possible to deliver medical packages from one base to another or to share drones between bases to save space.

\subsection{Maximum Distance Calculation}

Max payload capability $(MPC)$ mentioned in attachment 2 is redefined as the maxinum capability in which condition are in safety-guaranteed condition, and we defined an imaginary max drivable capability $MDC = k \cdot MPC$, in which condition are not able to fly, where $ k $ is the proportional coefficient.

Maximum flight time with cargo $ t $ is in negative relationship to cargo weight $ l $.Therefore, under any load $ l $, the maximum flight time can be derived as:

\begin{equation}
t = - \frac{ T } { MDC }  \cdot  l + T,  
\end{equation}
where T is the maximum flight time without cargo.

The power of flying to and back is different but time consumption of flying to and back is the same. In a round trip that approaches the farthest spot, total energy storage $E$ of a drone is used up:
\begin{equation}
  P_1  \cdot  t + P_2  \cdot  t = E \label{eq:P},
\end{equation}
where $P_1$ is the power of flying to, $P_2$ is the power of flying back, and $t$ is the time consumption of flying to or back:
\begin{equation}
t_1 = t_2 = \frac{D}{2V},
\end{equation}
where $D$ is the maximum reachable distance, and $V$ is the speed constant.
The power of flying to and back can be derived from division of energy and time:

\begin{equation}
P_1 = \frac{E}{t_1} = \frac{E}{ - \frac{ T } { MDC }  \cdot  l + T } 
\end{equation}
\begin{equation}
P_2 = \frac{E}{t_2}=\frac{E}{T} 
\end{equation}

Finally we subsituted equation (4,5) into equation (6) and derived maximum reachable distance $D$ under load $l$:
\begin{equation}
D = \frac{l -  k \cdot MCP}{l - 2k \cdot MCP}  \cdot  \frac{V \cdot T}{30} 
\end{equation}

\subsection{Cargo Bay Filling}

A genetic algorithm is used for this 3-D bin packing problem. We first enumerated all the combination of drones and package plans and then tested for feasibility of packing the package plan into the cargo bay. Here are the feasible drone package plans for each destination. Only top two plans with longest distance for each destination is shown.\label{genetic algorithm}

\begin{table}[htp] 
  \label{tbl Feasible drone plan with $k$ arbitrarily set as 1.3}
	\centering
	\caption{Feasible drone plan with $k$ arbitrarily set as 1.3}
% 	\resizebox{\linewidth}{
	\begin{tabular}{cccccccc}
		\toprule
		Destination& Drone& Bay Type& MED1& MED2&	MED3& Weight& Distance
		\\
		\midrule
		Caribbean Medical Center\\ Fajardo &	B &	1 &	1 &	0&	1 &	5 &	36
		\\
		Caribbean Medical Center\\ Fajardo&	C&	2&	1&	0&	1&	5&	31.39\\
		Hospital HIMA \\San Pablo&	C&	2&	2&	0&	1&	7&	28.44\\
		Hospital HIMA\\ San Pablo&	F&	2&	2&	0&	1&	7&	27.19\\
		Hospital Pavia Snturce\\ San Juan& B&	1&	1&	1&	0&	4&	40.13\\
		Hospital Pavia Snturce\\ San Juan& C&	2&	1&	1&	0&	4&	32.72\\
		Puerto Rico Children's Hospital\\ Bayamon&	F&	2&	2&	1&	2&	12&	23.21\\
		Puerto Rico Children's Hospital\\ Bayamon&	C&	2&	2&	1&	2&	12&	18.97\\
		Hospital Pavia Arecibo\\ Arecibo&	B&	1&	1&	0&	0&	2&	47.06\\
		Hospital Pavia Arecibo\\ Arecibo&	C&	2&	1&	0&	0&	2&	35.16\\
		\bottomrule
	\end{tabular}
% }
\end{table}

\subsection{Oversampling and K-means Clustering Model} 

Since the packing problem is simply organized, our focus is the location(s) of cargo container(s). We firstly oversampled the locations of hospitals due to the minority of hospitals compared with roads, and then we located the cargo containers by K-means to cluster.

Oversampling in data science is a technique, that is used to adjust the class distribution of a data set. The number of points from the road greatly exceeds the number of locations of hospitals, which leads to an imbalance in the data set. In details, if we give them the same weight to calculate the best locations, the result would certainly ignore the tiny turbulence of the locations of hospitals. Therefore, we decided to oversample the locations of hospitals by giving them more weights to emphasize the importance of hospitals. After oversampling, we used K-means clustering to classify three categories of hospitals and main roads, and find the K-means centers of three categories.

We conducted the algorithm of K-means clustering. Here, euclidean distance formula is defined as:
\begin{equation}
  dist(X, Y)=\sqrt{\sum_{i=1}^{n}\left|x_{i}-y_{i}\right|^{2}}
\end{equation}
Let $T=x_1, x_2, \cdots, x_m$ be the points to be clustered. The steps of k-means clustering are as follows:
\begin{enumerate}
  \item Initialize $k$ cluster centers $a_1,a_2,\cdots,a_k$;
  \item Assign each point to the cluster whose mean has the least squared Euclidean distance.
  \begin{equation}
    label_i = \mathop{\arg\min}_{1\leq j\leq k} {\sqrt{\sum_{i=1}^{n}(x_i-a_j)^2}}
  \end{equation}
  \item Recalculate the centroid of each cluster given as:
  \begin{equation}
    a_i = \left(\frac{\sum_
    {j}^{n} x_{j}}{n}, \frac{\sum_{j}^{n} y_{j}}{n}\right)
  \end{equation}
  \item Check if it meets a terminal condition, which could be evaluated by iteration times, mean squared error or cluster center varying frequency.
  \item Go to step 2.
\end{enumerate}

\begin{figure}[htb]
  \centering
  \includegraphics[width=0.5\textwidth]{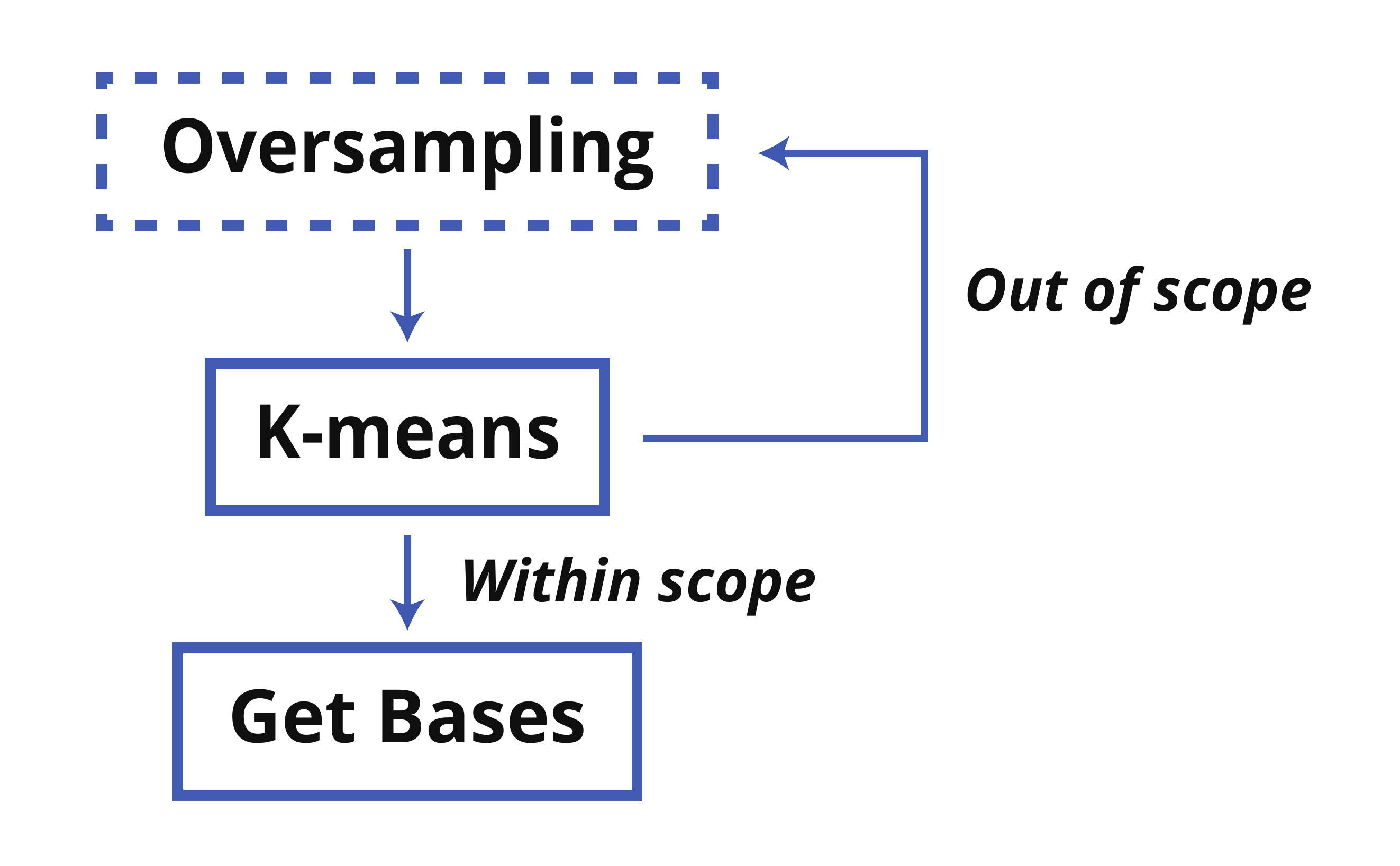}
  \caption{K-means}
  \label{fig:kmeans}
\end{figure}

In the realistic occasions, we see that the ratio of the number of hospitals and the number of sampled points on main roads are over 1:10, so we decided to oversample the number of hospitals into a ratio of 1:1. Then we put these points into K-means clustering, which would give back $k(1\leq k\leq 3)$ cluster centers back. To determine $k$, we prints three possible results of $k$. In the occasions of $k=1,2$, distances from cluster centers to hospitals are too long for drones to arrive by its limited batteries, according to Table 2. So we had to locate 3 cargo containers as the DroneGo disaster response system.

\begin{figure}[htb]
  \centering
  \includegraphics[width=0.8\textwidth]{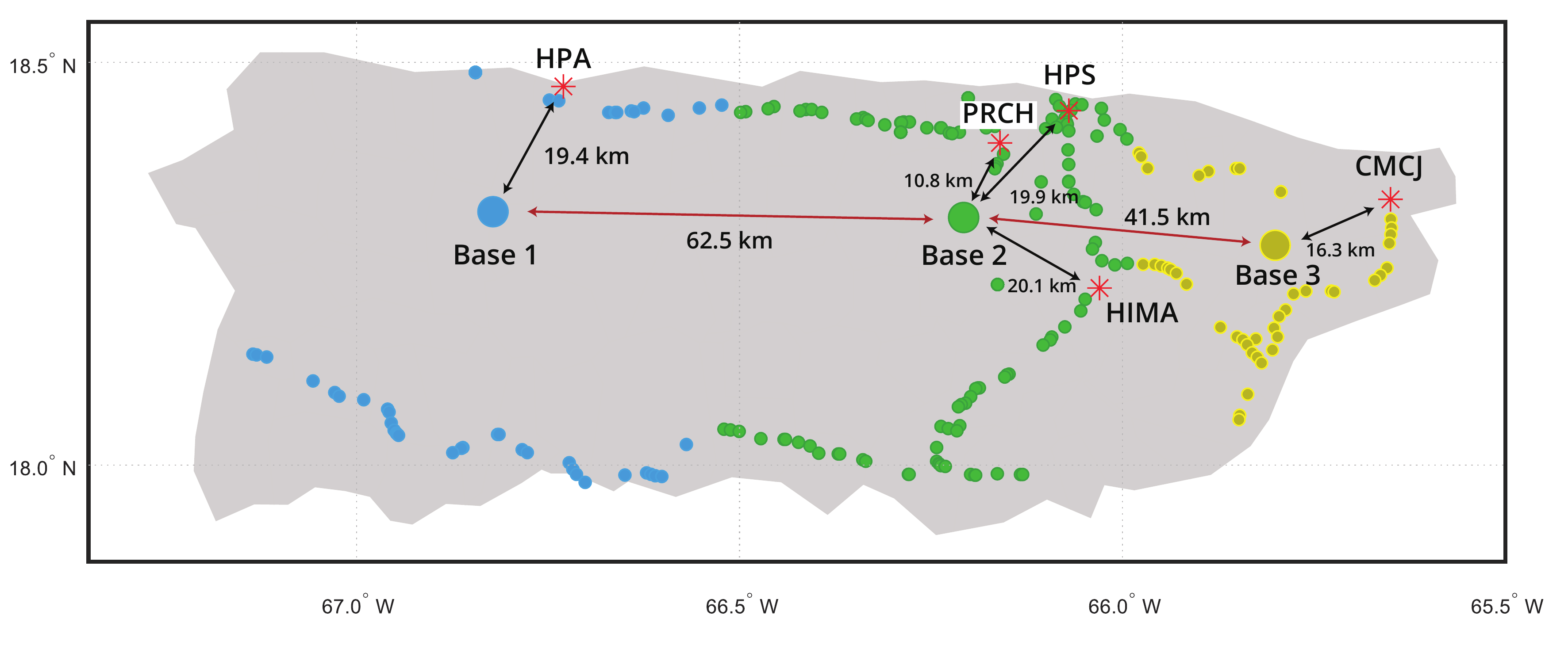}
  \caption{K-means Result}
  \label{fig:kmeansResult}
\end{figure}

The result of k-means clustering, which indicates the locations of cargo containers, is 
(18.3147,-66.8219), (18.298,-66.2065), (18.2698,-65.7574). The locations of cargo containers and sampled points on main roads are shown as figure \ref{fig:kmeansResult}.

Five destinations can be partitioned into 3 clusters: 
\begin{itemize}
\item Base 1: Caribbean Medical Center
\item Base 2: Hospital HIMA , Hospital Pavia Snturce and Puerto Rico Children's Hospital
\item Base 3: Hospital Pavia Arecibo
\end{itemize}

Base 1 and 3 have only one destination assigned and the optimal drone shown in table \ref{tbl Feasible drone plan with $k$ arbitrarily set as 1.3} for both bases are drone B. Thus drone B is our choice for base 1 and 3 with no doubt.

Base 2 has three destinations assigned, and the most optimal common drone of those three destinations is drone C (ranked first in Hospital HIMA, second in both Hospital Pavia Snturce and Puerto Rico Children’s Hospital). Therefore drone C should be chosen for delivery mission for base 2.

For reconnaissance, drone B is outstanding in terms of maximum flight distance (53 km), so each base should at least has one drone B for filming roads. Even though drones have service life more than 2 years, we still decided to double the drone numbers in case that accidents happen. Two drone H are also included for each base for communication purpose. 

Because the distance between base 2 and 3 is 41.5km and is lower than either the maximum reachable distance of drone B carrying MED1 or without loading, we considered transportation of medical packages between base 2 and 3. By moving all MED1 storage from base 2 to base3, we can even the medical package number ratio. The rest space of the container is filled with medical packages in the proportion specified by destination medicine demand.
The actual number of each medical package in each base is calculated by genetic algorithm metioned in \ref{genetic algorithm}. Our container configuration is shown as follows:

\begin{table}[htb]
  \centering
  \caption{Packing configuration for cargo containers}
  \label{tbl: Packing configuration for cargo containers}
  \resizebox{\textwidth}{!}{
      \begin{tabular}{cccccccc}
          \toprule
          Base & Distination & Drone Config & Drone of Delivery & MED 1 & MED 2 & MED 3 & Supporting Days \\
          \midrule
          1 & HPA & B$\times$4 H$\times$2 & B & 2251 & 0 & 0 & 375\\
          2 & HIMA HPS PRCH & B$\times$2 C$\times$2 H$\times$2 & C & 0 & 1080 & 2160 & 375 \\
          3 & CMC & B$\times$2 H$\times$2 & B & 1320 & 0 & 1320 & 1320  \\
          \bottomrule
      \end{tabular}
  }
\end{table}

\begin{figure}[htp]
  \caption{Drone playload packing configurations}
  \label{fig:package}
  \centering
  \subfigure[Drone B with MED1,3]{
      \label{fig:package11}
      \includegraphics[width=0.3\textwidth]{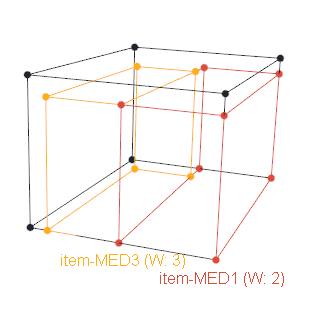}
  }
  \subfigure[Drone B with MED1,2]{
      \label{fig:package12}
      \includegraphics[width=0.3\textwidth]{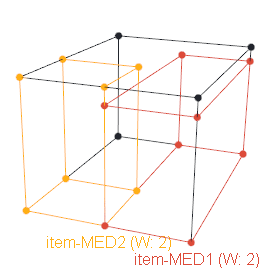}
  }
  \subfigure[Drone B with MED1]{
      \label{fig:package13}
      \includegraphics[width=0.3\textwidth]{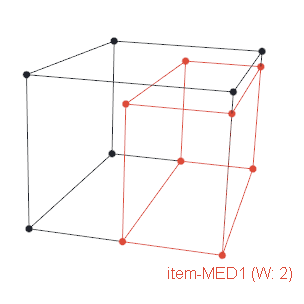}
  }      
  \medskip
  \subfigure[Drone C with MED1,1,3]{
      \label{fig:package21}
      \includegraphics[width=0.32\textwidth]{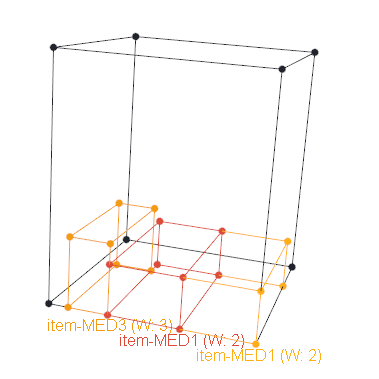}
  }
  \subfigure[Drone C with MED1,1,2,3,3]{
      \label{fig:package22}
      \includegraphics[width=0.32\textwidth]{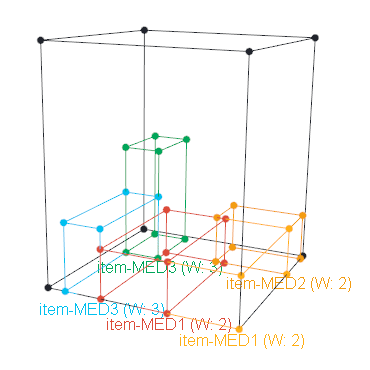}
  } 
\end{figure}

We visualized the packing configuration of each package plan (figure \ref{fig:package}). 

\section{Part2: Route designing by Biased Random Walk Model}
\subsection{Analysis of Problem}

In previous section, we decided the drone type for each base. All three bases have drone B, which can also fly farther than any other drones and has the video capability. Therefore, it's our perfect choice for reconnaissance. 

When we thought about the problem, we first come up with a deterministic model of path searching. However, deterministic model has its own flaws: 
\begin{itemize}
  \item It's easy to be trapped in a regional optima. 
  \item It depends too much on some arbitrary parameters.
  \item It only gives one feasible route. If accidents happen, namely wildfire, and block some cells, the entire route is not applicable anymore.
  \item It can not amend the route automatically when trapped in dead ends.
\end{itemize}

Thus, we decided to build a stochastic model simulating a random walk of "drunk man" with the following advantages:
\begin{itemize}
  \item It's expected to explore all feasible routes and thus find the global optima for us.
  \item It's naturally compatible to stochastic parameters. Thus, our biased walk is less biased by human discretion and more affected by the environment.
  \item It gives many feasible routes, many of which are equally good. The alternative routes are good backup routes.
  \item We don't have to consider how to design a feasible route because we can easily abandon infeasible routes.
\end{itemize}

We use the metaphor of a drunk man to blueprint our model. Our drunk man should behave as follows:
\begin{enumerate}
  \item Biased: attracted by roads;
  \item Exploratory and nostalgic: moving apart from the origin in the beginning and come back home in the end;\item New-trumps-old mindset: reluctant to walk on roads that it has walked on previously.
\end{enumerate}

Accordingly, our drunk man should sense the following factors:\label{factors}
\begin{enumerate}
  \item Importance: evaluated by road type.
  \item Altitude change: drones fly at the same height from ground.
  \item Moving distance: drones should go back to base before it's running out of power.
\end{enumerate}

After sensing the necessary factors, our drunk man can finally take the first step! But it still needs to integrate all those factors. But it's in its mind, anyway. Let's see how it works.

\subsection{Preliminary Model}
\subsubsection{Random Walk Procedure}

Imagine on Christmas eve a man gets loaded for some reason. It has nothing to do. Why not take a walk? So it decides to wander around the field. Departing from home, sometimes it would explore off the roads, but as a sane man, it would keep speeding as much time as possible on roads. Here is what's going on in its mind: First, it looks around and randomly pick a place to step on. It's not a psychopath at least so it would avoid hitting a tree or fall into rivers. After spotting a place, it would stare at it for a while to evaluate the spot it chose. Then it accept or reject the spot based on its evaluation. Keep in mind it's drunk so it's likely that it would be willing to accept a spot deviating from the main road. But the man has a good memory even when unconscious, so it would avoid going to the place once again (but is not compulsory). After moving to a new spot, it repeats the previous steps.

In parallel universes, sometimes our drunk man would fall into some trenchs (scenario 1); sometime it would get exhausted (scenario 2); but sometimes the man finally get home (scenario 3)! That's a good news, and the track of its walk is what we are looking for. 

It's time to implement our drunk man in algorithm. First of all, we initalized the drone sited on our bases. The drone keeps iterating the following steps on the field till one of three scenarios mentioned above happens.

\begin{enumerate}
  \item Search for all available adjacent cells, generate a random number and propose one of them according to a uniform probability;
  \item Evaluate the proposed cell by an objective function; which should gives a value between 0 to 1;
  \item Decide whether to accept or reject the proposal. Generate another random number for making decision. If lower than the value in step 2, it accepts the proposal, decrease the importance of the adjacent cells and moves to the proposed cell. If it's lower, it rejects the proposal and keeps still.
\end{enumerate}

\subsubsection{Objective Funcion}

The objective function integrates all information necessary to our model and is the only source affecting  drones' tendency of filming road networks or going home. It evaluate the relative superiority of proposed cell 
to other adjacent feasible cells. 

We supposed three elements, $p_{road}$, which affects the tendency of drones flying toward roads, $p_{home}$, which affects the tendency of drones going home, and $\gamma$, which is the relative weight of $p_{home}$. The value $f$ of objective function is given as:
\begin{equation}
  f = p_{road} + \gamma \times p_{home}
\end{equation}

Let $p_{road} = \frac{road_i}{\sum road_i}$, and $p_{home} = \frac{home_i}{\sum home_i}$, where $road_i$ means the significance of selected point by K-nearest neighbor rule (described later), and $home_i$ means the distance from the selected point to the origin. Here, $p_{road}$ and $p_{home}$ are nondimensionalized by other feasible cells adjacent to the current cell, so that $\sum p_{road}=\sum p_{home} = 1$, and therefore $p_{road}$ + $p_{home}$ is plausible. We enlarge the difference between $home_i$ and $road_i$ by exponetialization. Thus, let
$$\hat{p}_{road} = \frac{e^{road_i}}{\sum e^{road_i}},\quad \hat{p}_{home}=\frac{e^{home_i}}{\sum e^{home_i}}$$
Still, $\hat{p}_{road}$ and $\hat{p}_{home}$ are nondimensionalized\footnote{Detailed calculation is omitted. For real implementation, please check our code in appendix}.

To make our drone exploratory and nostalgic, we defined the relative weight of home attraction $\gamma = \alpha \times (d/MFD - \beta )^3 $ where $\alpha = 0.2 \& \beta = 0.3$ are set arbitrarily, $d$ is the distance from current cell to the origin and $MFD$ is maximum flight distance, $MFD = 53000$ for drone B. $\gamma$ is negative at the beginning of the route, making cells apart from the origin have higher probability to be accepted. When leaving origin for some distance, $\gamma$ is close to zero and $p_{road}$ takes major place in accpeting or rejecting proposed cells. In this period the drone behaves totally based on road attraction. In final stage when flight distance is approaching $MFD$, $p_{home}$ is in charge of the process and the drone would accept proposed cells directing to the origin with higher probability.

The sensing range of the drone is specified by the K-nearest neighbor rule. When larger neighbor rule  is used, the drone is able to be attracted to remote roads if the drone is wandering within a non-road space. However, if roads are more dense in the field, it's possible that the heterogeneity of the field is blurred out and thus the drone that moves along roads in small rule would move unintentionally. In this paper, Four|eight|twelve-nearest neighbor rules are implemented in our program and the results are discussed in sensitive analysis.

\begin{figure}[htb]
  \centering
  \subfigure[Four-nearest neighbor rules]{
    \includegraphics[width=0.3\textwidth]{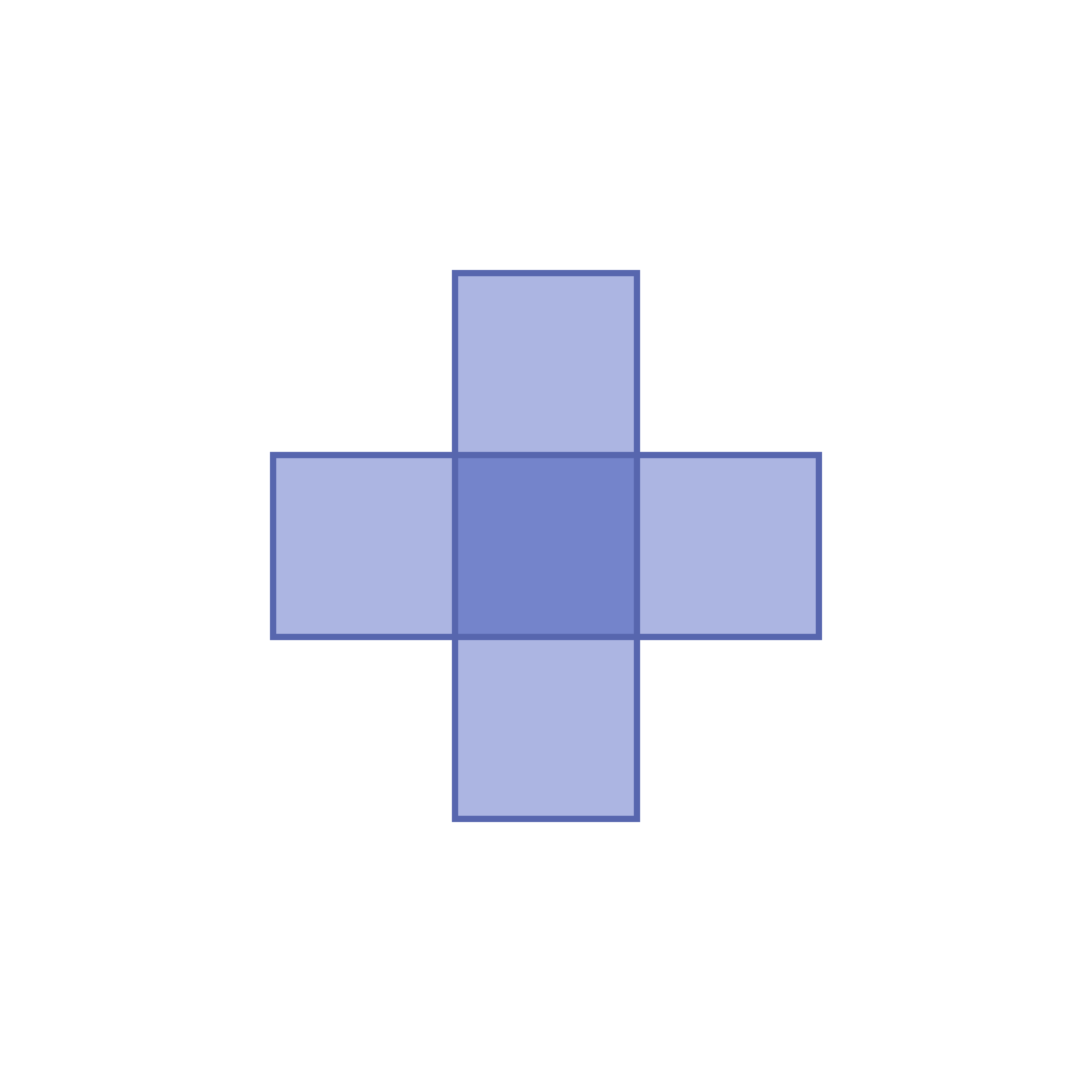}
  }
  \subfigure[Eight-nearest neighbor rules]{
    \includegraphics[width=0.3\textwidth]{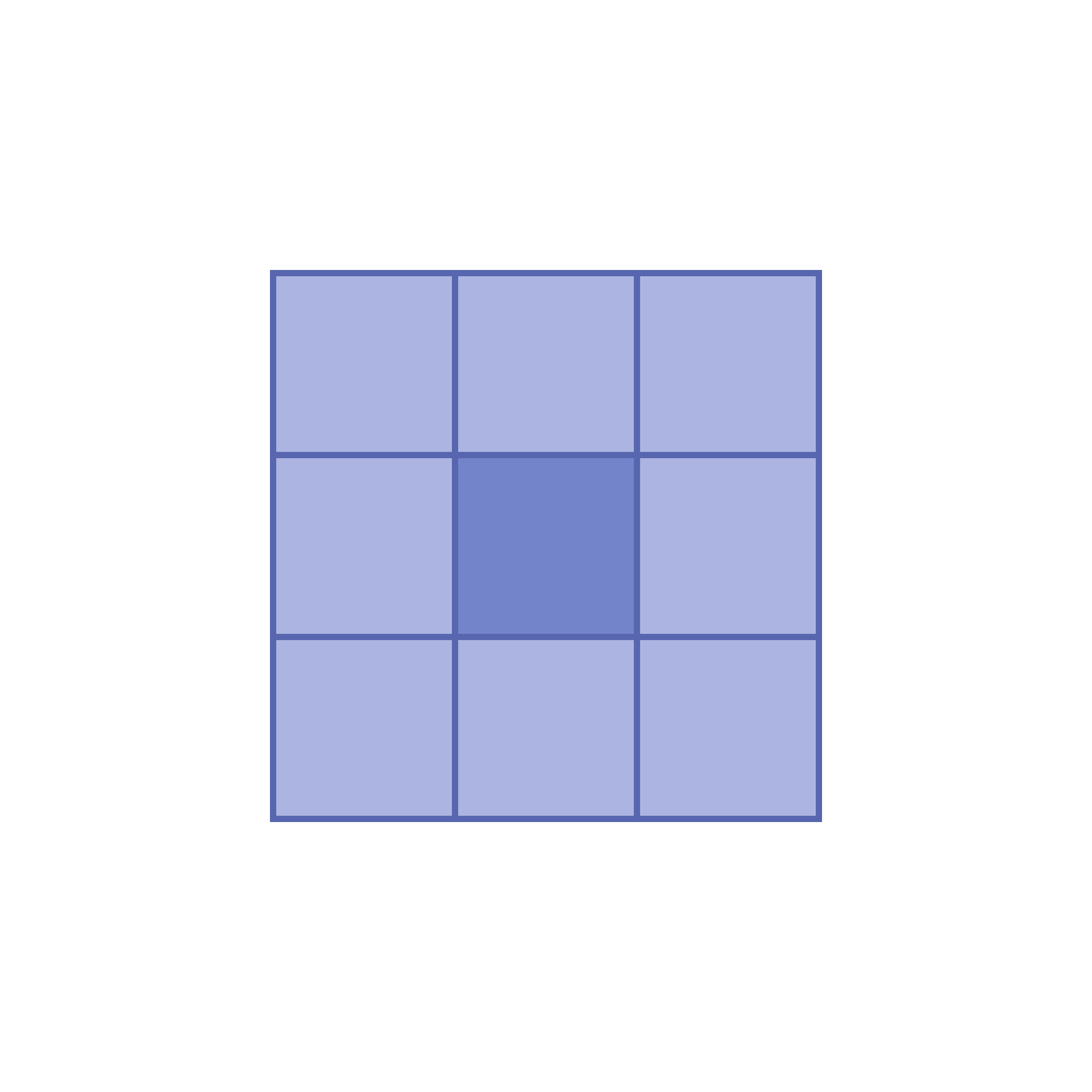}
  }
  \subfigure[Twelve-nearest neighbor rules]{
    \includegraphics[width=0.3\textwidth]{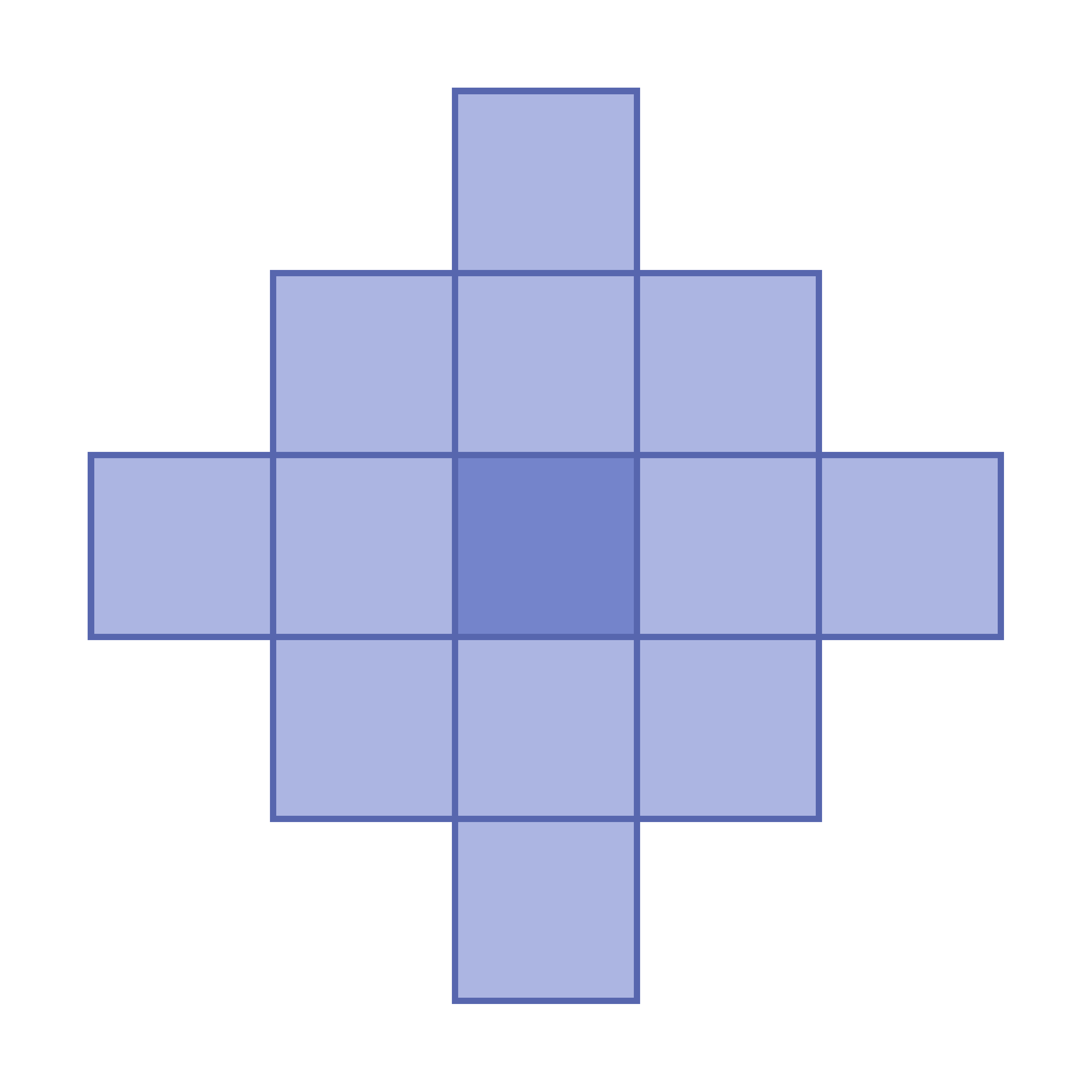}
  }
  \caption{K-nearest neighbor rule}
  \label{fig:neighbor}
\end{figure}

\subsubsection{Results of Preliminary Model}

\begin{figure}
  \centering
  \caption{The results of preliminary model}
  \label{fig:routepm} 
  \subfigure[Route 1]{
      \label{fig:route43}
      \includegraphics[width=0.45\textwidth]{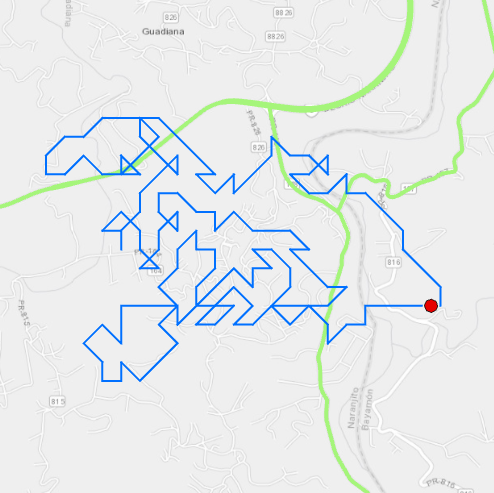}
  }
  \subfigure[Route 2]{
      \label{fig:route139}
      \includegraphics[width=0.45\textwidth]{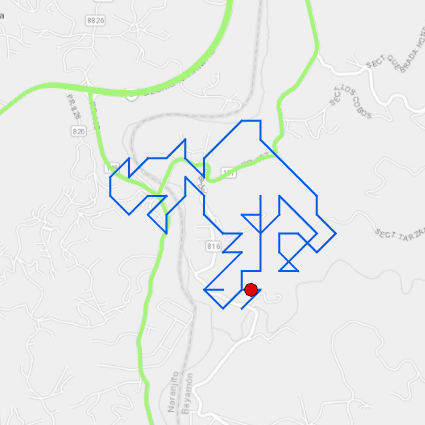}
  }
  \end{figure}

After letting off 10000 drones, we picked the route with the highest coverage and ploted in \ref{fig:routepm}.  As we can see, in fig a, the drone spent too much time off roads and wandering on the left-bottom corner of the picture. It depicts that roads are not attracted by roads to enough extent. In fig b, the drone covers some part of the nearest road but fail to move farther to explore more territory. Therefore, we concluded $\gamma$ is too small in both ends. Besides, the field is discontinuous at most parts, therefore unbiased walk is commonly expected when drones are trapped in a non-road field. Therefore, some irrational looping is inevitable.

\subsection{Modified model}
\subsubsection{Unleashing Parameters}

In defining the relative weight of home attraction $\gamma$, two nuisance parameters $\alpha \& \beta$ are used: $$\gamma = \alpha \times (d/MFD - \beta )^3. $$ $\beta$ determines when $\gamma = 0$ and thus pinpoints the turning point of drones from exploratory (going out) to nostalgic (going back). After rescaled by $\alpha$, the $\gamma$ is increased or decreased, making home attraction more or less important in proposal mechanism.

We modified our model by unleashing those two parameters from our hand. We set $\alpha$ and $\beta$ to follow a log normal distribution with means equal $0.5$ and $0.5$ respectively and variance equal $0.7$ and $0.05$ respectively. Those numbers are derived arbitrarily. According to previous analysis, $\gamma$ is small in both direction, we elevated the mean of scalar $\alpha$. After we simulated a feasible route, we would take down its $\alpha \& \beta$ and distance for further analysis. 

\subsubsection{Experience Learning}

Importance data is naturally discrete but experience of previous drones is not. If we use Convolutional Neural Networks(CNN), it cannot memorize the former choices. In order to fit our biased random walk model, we should set a neural network that can memorize the former choices and decide its next move with the turbulance of former choices. 

Here we introduce Recurrent Neural Network(RNN). RNN can make use of a sequential information, which means that RNN can memorize former choices and decide its next move by some kind of weight of former choices. In our case, according to our Biased Random Walk Model, we choose the furthest distance that drones go, the repeated routes that drones have, and the standard deviation as our parameters into RNN, trying to train more stable and more accurate parameters into the previous model.

\begin{figure}[htb]
  \centering
  \includegraphics[width=0.8\linewidth]{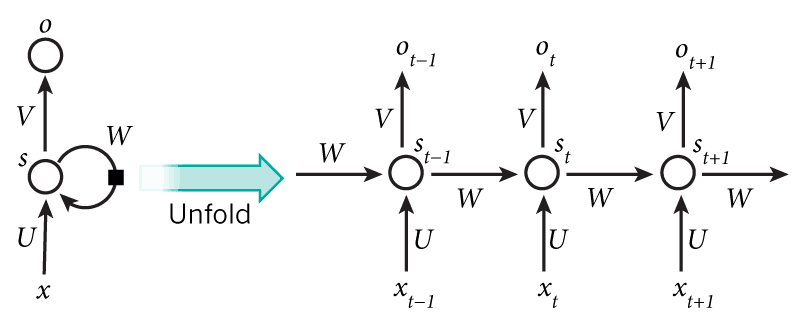}
  \caption{The principle of the recurrent neural network\cite{lecun2015deep}}
  \label{fig:rnn}
\end{figure}

The figure \ref{fig:rnn} is the unfolded picture of Hidden Layer of RNN. $t-1, t, t+1$ depicts a Time-Series set, X means the input data set, $S_t$ means the memory of sample situated in the time $t$, $S_t = f(W*S_{t-1} + U*X_t)$, where $W$ means the input weight, $U$ means the weight of sample in this time, $V$ means the output weight.

When $t=1$, we initialize $S_0=0$, randomly initialize $W, U, V$, and calculated by the equations below:
$$h_1=U \dot x_1+W\dot s_0$$
$$s_1=f(h_1)$$
$$o_1=g(Vs_1)$$
where function f and function g are both activation function, f can be classic tanh, relu and sigmond function, g can be softmax function. 
When $t=2$,
$$h_2=U \dot x_2+W\dot s_1$$
$$s_2=f(h_2)$$
$$o_2=g(Vs_2)$$
As shown above, we can deduce the general equation from this:
$$h_t=U \dot x_t+W\dot s_{t-1}$$
$$s_t=f(h_t)$$
$$o_t=g(Vs_t)$$

In our Random Walk model, we remembered the former state and then made its next move. Training the model by RNN, can enhance performance of our model.

  \begin{figure}
    \centering
    \caption{The results of modified model}
      \label{fig:routemm}
      \subfigure[Route 3]{
      \label{fig:route25035}
      \includegraphics[width=0.45\textwidth]{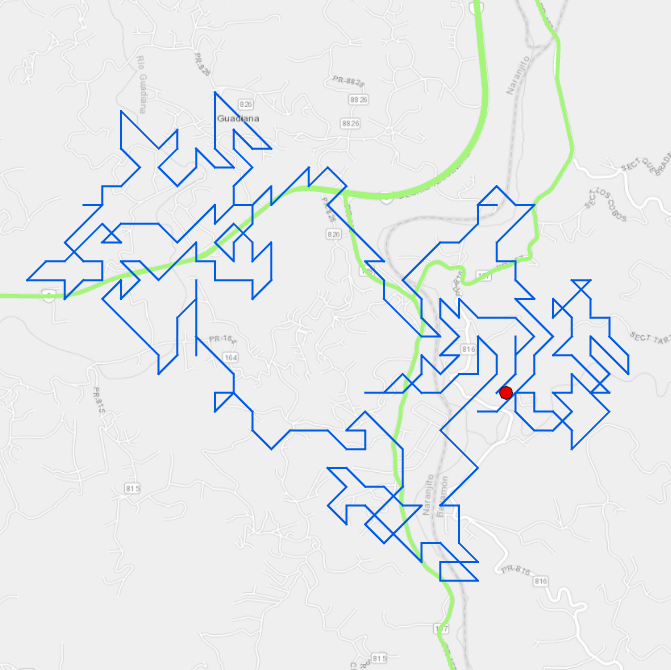}
      }
      \subfigure[Route 4]{
          \label{fig:route25097}
          \includegraphics[width=0.45\textwidth]{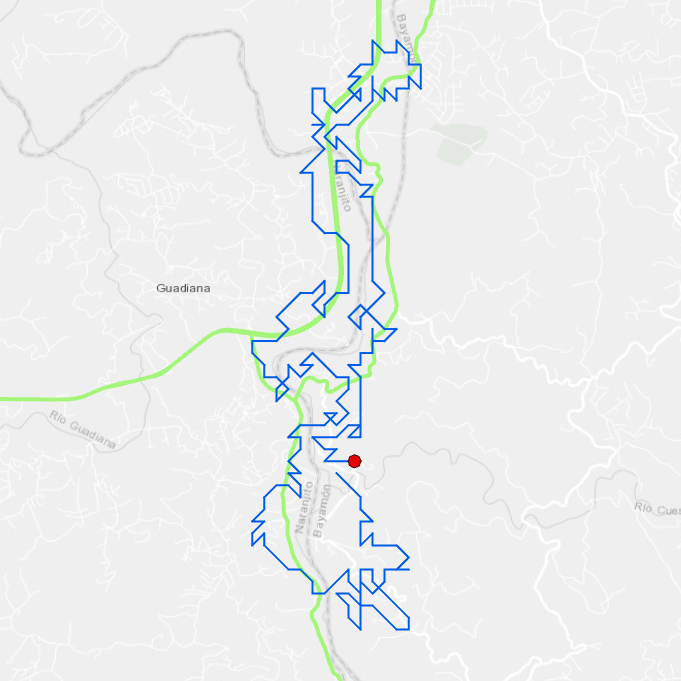}
      }        
      \subfigure[Route 5]{
          \label{fig:route66711}
          \includegraphics[width=0.45\textwidth]{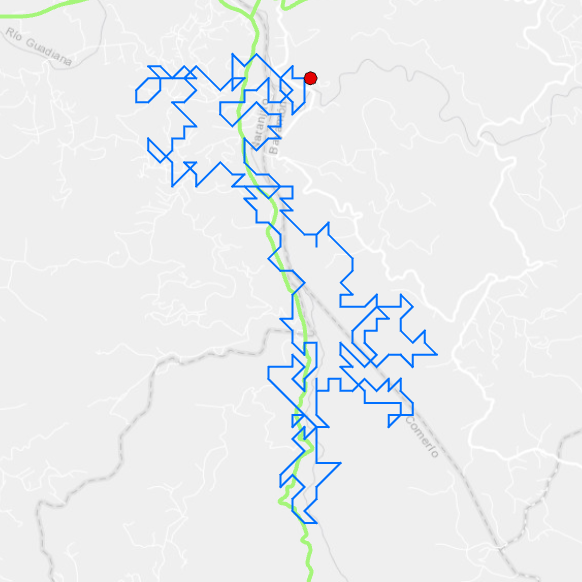}
      }
      \subfigure[Route 6]{
          \label{fig:route75036}
          \includegraphics[width=0.45\textwidth]{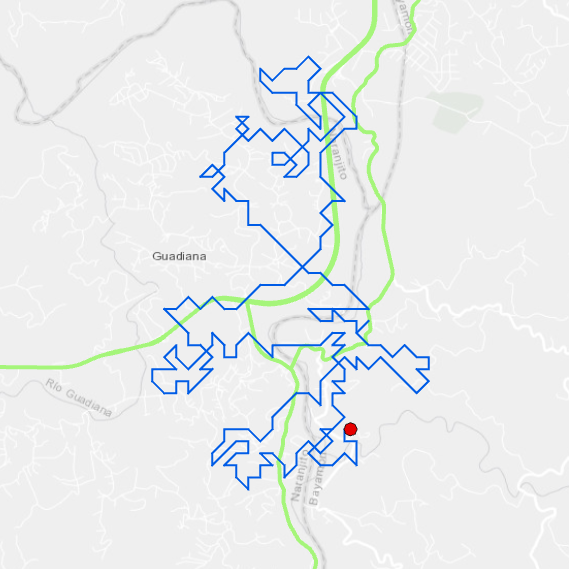}
      }
\end{figure}

\subsection{Results of Modified Model}

The results of modified model are plotted in fig \ref{fig:routemm} from a to d. As we can see, those routes are significantly better than results of preliminary model. For example, the drone of route in figure \ref{fig:route25035} walked along roads for a long distance and crossed the non-road field as if roads on the right botton corner are pulling it. However, some small loops  still exists in all four routes. 

For better coverage, we tried to overlay 2 routes chosen from all feasible routes. Conbinatorial Coverage Rate(CCR) is derived from $ CCR = \frac{NC}{A} $, where $NC$ is the net coverage defined as the coverage of  union of two routes, and $A$ is the area of a specified bounding box.

Combination of route in figure \ref{fig:route25035} and \ref{fig:route25097} has the highest conbinatorial coverage rate (0.82) among all possible plans.

\section{Sensitivity Analysis of Model}
\subsection{Distribution Change of Behavior Parameters}

In modified model, we unleashed behavior parameters $\alpha$ and $\beta$ to follow a lognormal distribution. If we only save the value of those two parameters of routes whose coverage surpasses some threashold, our model could be seen as a gate that filters out some paramters that are not "fittable" intuitively for (1) generate a feasible route (2) increase the coverage. 

\begin{figure}[htb]
  \centering
  \caption{Regression analysis of $\alpha$ and $\beta$}
  \label{fig:coverage}
  \subfigure[Alpha]{
    \includegraphics[width=0.45\textwidth]{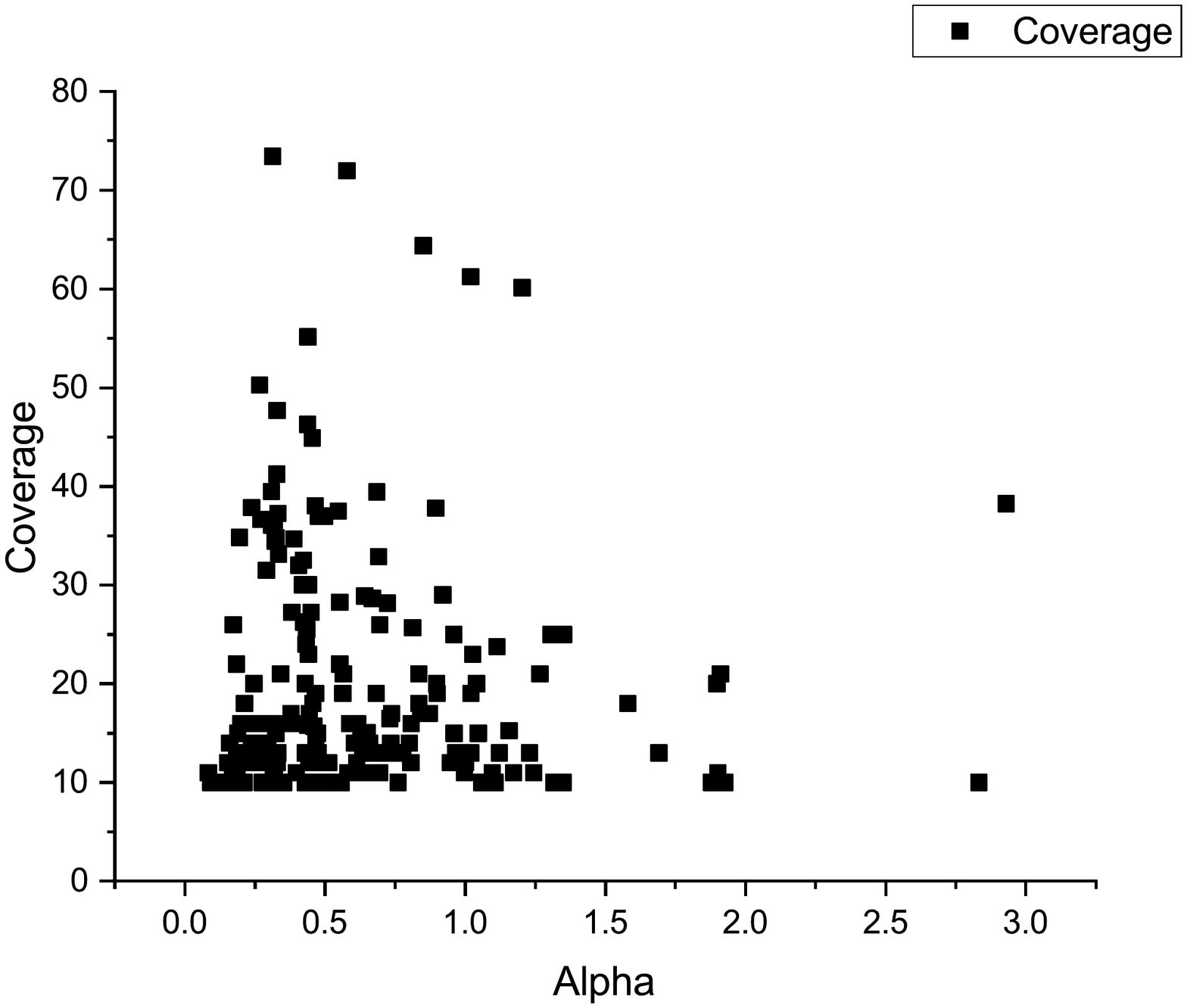}
    
  }
  \subfigure[Beta]{
    \includegraphics[width=0.45\textwidth]{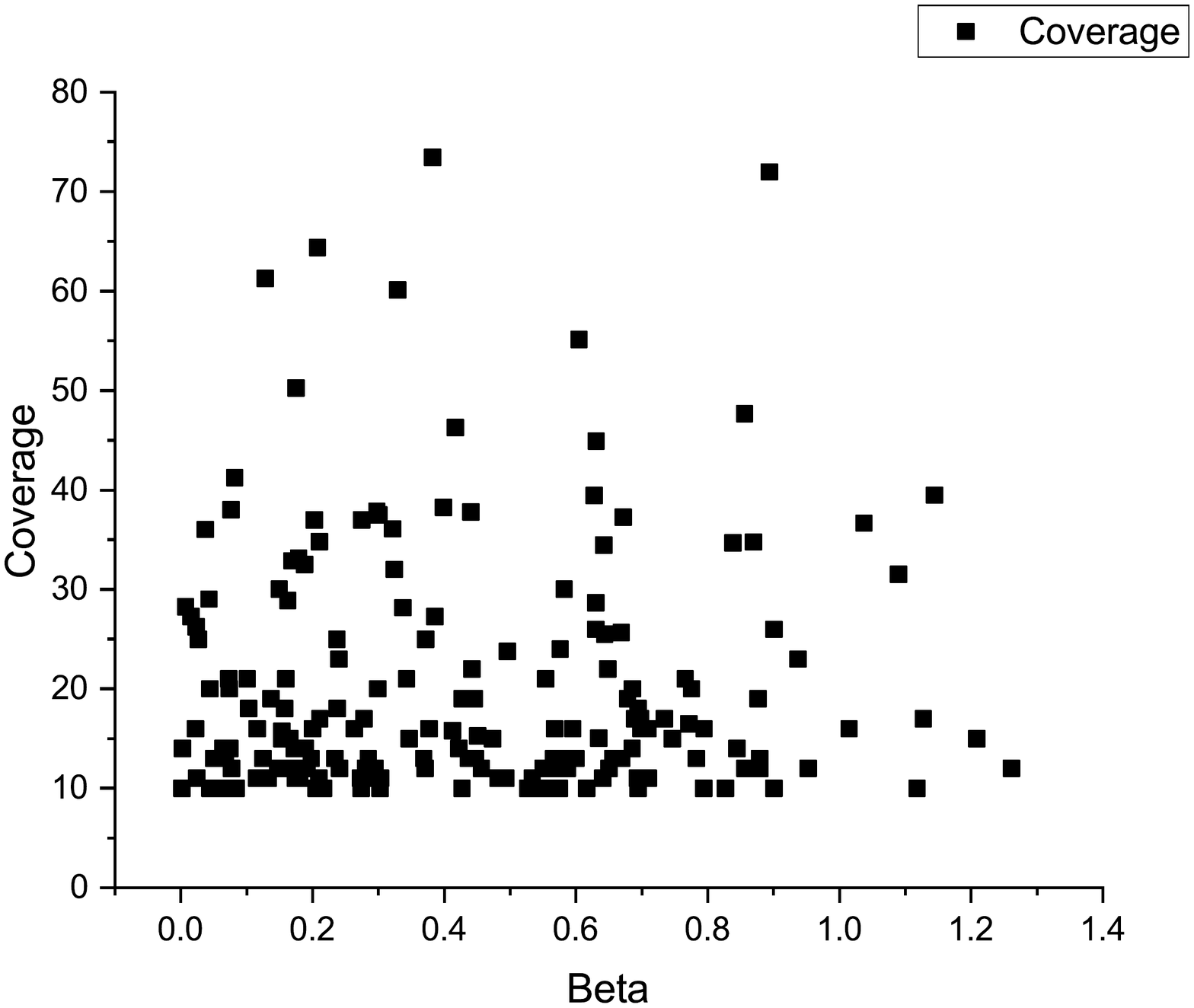}
    
  }
\end{figure}

\begin{figure}[htb]
  \centering  
  \caption{Lognormal fitting of $\alpha$ and $\beta$}
  \label{fig:logfit}
  \subfigure[Alpha]{
    \includegraphics[width=0.45\textwidth]{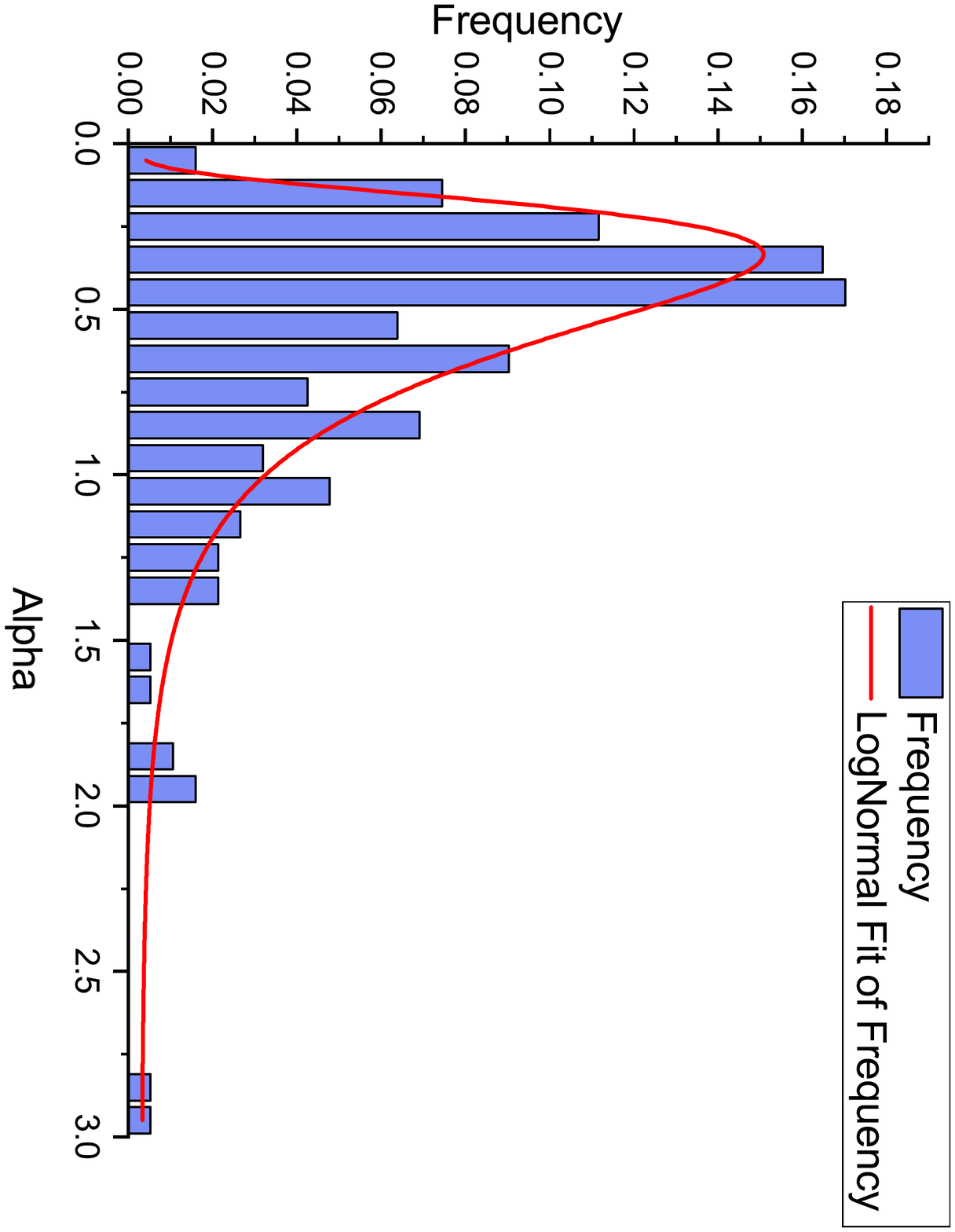}
            
  }          
  \subfigure[Beta]{
    \includegraphics[width=0.45\textwidth]{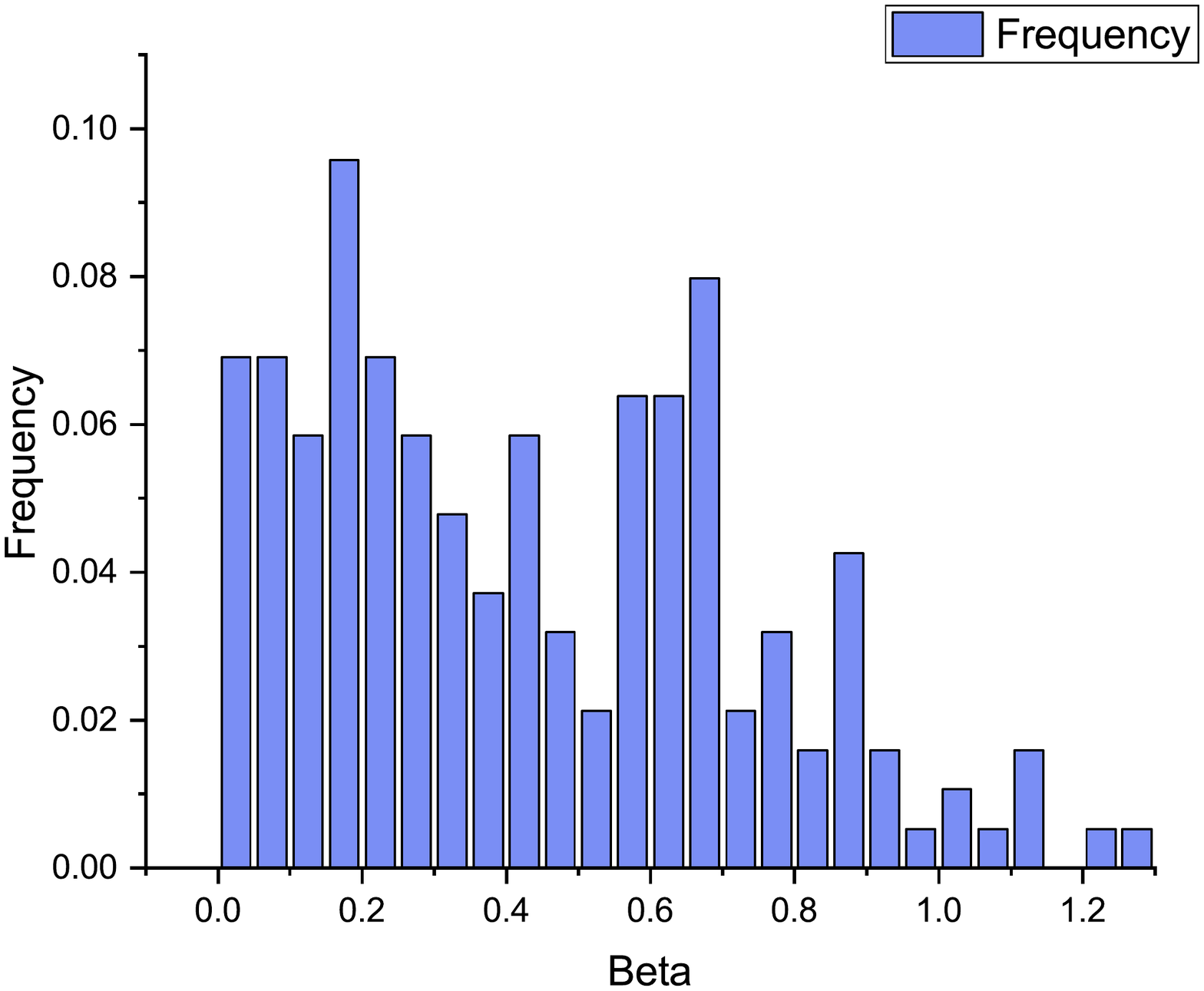}
    
  }
\end{figure}

First, we did a regression analysis of $\alpha$ and $\beta$ to coverage. However, there is no significant relationship between any two of them (figure \ref{fig:coverage}). Second, we did a lognormal fitting of $\alpha$ and $\beta$ respectively (figure \ref{fig:logfit}). 
$\alpha$ still fits a lognormal distribution($R^2 = 0.89$), with consistant mean value and a small deviation of variance from $0.7 \to 0.61$. In terms of $\beta$, it fails to fit a lognormal distribution and has the mean value $ 0.43 $ and variance $0.29$.

% coverage

% frequency fit

This analysis shows that neither $\alpha$ or $\beta$ has contribution to the performance of the biased random walk process. The distribution of $\alpha$ is not significantly affected after filtering. However, the distribution of $\beta$ is evened. 
\begin{equation}
  \hat{f}(x)=\left\{
  \begin{aligned}
  g(x)<f(x), x \in K \\
  f(x), x \notin K
  \end{aligned}
  \right. K \in U(\mu, \epsilon);
  \end{equation}
So, we can get that,
$$M[\hat{f}(x)]=M_{x\in K}[\hat{f}(x)]+M_{x\notin K}[\hat{f}(x)]=M_{x\in K}[g(x)]+M_{x\notin K}[f(x)]\leq M[f(x)]$$
which means that the probability of after filtering is smaller than the prob  

$\beta$ has explicit meaning in our model --- time when home attraction surpasses road attraction. Intuitively we expect our drone to turning from departing to going back in the halfway. However, this expectation is probably flawed according to our result. A drone either tending to move near its origin or go far away is probably more likely to make a feasible route and has higher coverage. It occurred to our mind that maybe the lognormal distribution is not a good distribution for our nuisance parameters. And even distribution would generate a more interpretable distribution for results because it's not skewed at the beginning. However, even distribution is not an uninformative distribution\cite{gelman2013bayesian}, and thus it will still affect the result.

\subsection{Performace Change due to K-nearest neighbor rule}

We conducted our model based on different K-nearest neighbor rule. The 8-nearest neighbor rule performs best according to average distance and coverage, and 4-nearest neighbor rule worst. The relationship between rule and performance is not deterministic. Field heterogeneity plays a key role. Therefore, we cannot interpret anything from results of a single field. More artificial field should be generated for further analysis.

\section{Stength and Weakness}
\subsection{Strength}
\begin{enumerate}
  \item We used real data for analysis and correctly projected points based on coordinate reference system.
  \item We used stochastic model to generate feasible routes to minimize anthropogenic bias  underlying evaluation method.
  \item The model is space-explicit --- it uses as much spatial data as possible to reflect the reality. No simplication on data side is applied apart from importance assignment of roads.
  \item This model is flexible that many parts could be modified to integrate new factors biasing our route designing.

\end{enumerate}
\subsection{Weakness}
\begin{enumerate}
  \item We have no idea how to tradeoff between delivery distance and road coverage so we arbitrarily emphasized the latter, which may be seen as penny wise and pound foolish.
  \item We simulate one drone at a time. Therefore, the combination of routes would have too many overlapping parts.
  \item The behavior of our drone is uncontrollable. For example, local looping of routes are inevitable. Afterwards smoothing is required.
  \item The algorithm is time-consuming and even though does not guarantee to find the best solution.
  \item The initial distribution of nuisance paramters still has some arbitrariness, and filtered distribution is not interpretable.
\end{enumerate}

\bibliographystyle{unsrt}  


\begin{thebibliography}{1}

\bibitem{owidnaturaldisasters}
Hannah Ritchie and Max Roser.
\newblock Our World in Data, Natural Disasters.
\newblock In {\em https://ourworldindata.org/natural-disasters}, 2019.

\bibitem{OpenStreetMap}
OpenStreetMap contributors.
\newblock Planet dump retrieved from https://planet.osm.org.
\newblock In {\em https://www.openstreetmap.org}, 2019.

\bibitem{altitudedata}
Geological Survey (U.S.)
\newblock 100-Meter Resolution Elevation of Puerto Rico and the U.S. Virgin Islands, Albers projection
\newblock In {\em National Atlas of the United States}, 2010.

\bibitem{li2014genetic}
Li, Xueping and Zhao, Zhaoxia and Zhang, Kaike
\newblock A genetic algorithm for the three-dimensional bin packing problem with heterogeneous bins
\newblock In {\em IIE Annual Conference. Proceedings, Page 2039, Institute of Industrial and Systems Engineers (IISE)}, 2014.

\bibitem{LSTMNetworks}
Christopher Olah
\newblock Understanding LSTM Networks
\newblock In {\em http://colah.github.io/posts/2015-08-Understanding-LSTMs/}, 2015.

\bibitem{martello2000three}
Martello, Silvano and Pisinger, David and Vigo, Daniele
\newblock The three-dimensional bin packing problem
\newblock In {\em Operations Research, Page 256--267}, 2000.

\bibitem{themistocleous2014use}
Themistocleous, Kyriacos
\newblock The use of UAV platforms for remote sensing applications: case studies in Cyprus
\newblock In {\em Second International Conference on Remote Sensing and Geoinformation of the Environment (RSCy2014), International Society for Optics and Photonics}, 2014.

\bibitem{wu2010three}
Wu, Yong and Li, Wenkai and Goh, Mark and de Souza, Robert
\newblock Three-dimensional bin packing problem with variable bin height
\newblock In {\em European journal of operational research}, 2010.

\bibitem{daniel2009airshield}
Daniel, Kai and Dusza, Bjoern and Lewandowski, Andreas and Wietfeld, Christian
\newblock AirShield: A system-of-systems MUAV remote sensing architecture for disaster response
\newblock In {\em Systems conference, 2009 3rd Annual IEEE}, 2009.

\bibitem{yu2003extensions}
Yu, Chaoqing and Lee, JAY and Munro-Stasiuk, Mandy J
\newblock Extensions to least-cost path algorithms for roadway planning
\newblock In {\em International Journal of Geographical Information Science}, 2003.

\bibitem{kitjacharoenchai2019multiple}
Kitjacharoenchai, Patchara and Ventresca, Mario and Moshref-Javadi, Mohammad and Lee, Seokcheon and Tanchoco, Jose MA and Brunese, Patrick A
\newblock Multiple Traveling Salesman Problem with Drones: Mathematical model and heuristic approach
\newblock In {\em Computers \& Industrial Engineering}, 2019.

\bibitem{saha2005gis}
Saha, Ashis Kumar and Arora, Manoj K and Gupta, Ravi Prakash and Virdi, ML and Csaplovics, Elmar
\newblock GIS-based route planning in landslide-prone areas
\newblock In {\em International Journal of Geographical Information Science}, 2005.

\bibitem{lecun2015deep}
LeCun, Yann and Bengio, Yoshua and Hinton, Geoffrey
\newblock Deep learning
\newblock In {\em Nature Publishing Group}, 2015.

\bibitem{gelman2013bayesian}
Gelman, Andrew and Stern, Hal S and Carlin, John B and Dunson, David B and Vehtari, Aki and Rubin, Donald B
\newblock Bayesian data analysis
\newblock In {\em Chapman and Hall/CRC}, 2013.


\end{thebibliography}

\end{document}